\documentclass[11pt]{article}
\usepackage[letterpaper, margin=1in]{geometry}
\usepackage[utf8]{inputenc}
\usepackage[T1]{fontenc}
\usepackage{times}
\usepackage{amsmath}
\usepackage{amsfonts}
\usepackage{amsthm}
\usepackage{nicefrac}

\usepackage{graphicx}
\usepackage{booktabs}
\usepackage{wrapfig}
\usepackage{xcolor}
\usepackage{natbib}
\usepackage{microtype}
\usepackage{enumitem}
\setlist{leftmargin=*, itemsep=0pt, topsep=0pt, parsep=0pt, partopsep=0pt}
\usepackage{url}
\usepackage[colorlinks=true, linkcolor=black, citecolor=black,
            urlcolor=blue]{hyperref}

\title{The Authority Expectancy Effect in Multi‑Party Conflict
}

\date{}

\author{
  Eunna Lee \\
  Independent Researcher \\
  \texttt{eunna.lee.ai@gmail.com}
}

\begin{document}

\maketitle

\begin{abstract}
In multi-party competitive settings, across experiments on resource allocation, fault attribution, and dispute mediation, we ask whether social authority (SA) cues, including occupational standing and institutional documentation, behave as if a scalar weight were added to one side of a judgment. We make this scalar-weight model explicit as a null hypothesis and reject both of its predictions. Its first prediction, that the effect of one cue does not depend on the presence of another, fails at the level of allocation: fitted to all runs, the occupation-by-documentation interaction is significant in two of the five models that could be fitted, with opposite signs, and in one model the reversal occurs without a single refusal, placing it entirely in the choice itself. Its second prediction, that an evidentiary cue shifts judgment in a fixed direction, fails in a multi-turn dispute: identical documentation protects the lower-authority party when that party holds it, but reverses or dissolves that protection when the higher-authority party does. The cues additionally govern whether a judgment is issued at all: removing an occupational label while retaining documentary evidence drives four of six models past both single-cue baselines into refusal, yet in two models that same label sharply reduces refusal when documentation is present, so the gate is subject to the same interaction as the judgment itself. We formalize the resulting pattern as the Authority Expectancy Effect (AEE) and characterize it through two properties: \textit{evidential reinterpretation}, whereby identical content is treated differently in the resulting judgment depending on which party bears the SA signal, and \textit{direction sensitivity}, whereby the same evidentiary cue moves outcomes in opposite directions depending on whether it aligns with the holder's authority position.
\end{abstract}

\section{Introduction}
\label{sec:introduction}
Vulnerability signals are typically given priority in order to protect the most severely harmed parties—a principle that is reflected in contemporary LLM alignment objectives for harm avoidance. Individuals, however, also carry social attributes—relative social standing or positional influence—that function as social authority (SA) signals, forming a second axis that can reinforce or counteract vulnerability-based prioritization. When mediating disputes, model outputs may therefore vary with the severity implied by the case and with the social cues attached to each party, with the interaction between these two dimensions shaping the resulting judgments.

We term this interaction the \textit{Authority Expectancy Effect} (AEE): the effect of an SA cue depends on the social configuration it appears in, so the same evidence does not shift outcomes by a fixed magnitude and direction. AEE is not reducible to authority bias, under which higher-status claims simply receive greater credibility. Occupational labels operate in distinct ways: supplying a causal narrative that assigns fault to one party, and reversing protective allocation depending on how authority aligns with documentary evidence. These effects vary by model and by SA variable, producing divergent trajectories that additive reweighting does not explain. 

\paragraph{Contributions.} First, we formalize the scalar-weight model implicit in prior work and specify two falsifiable predictions, providing a null against which authority effects can be tested rather than described (\S\ref{sec:concept}). Second, we treat refusal as an outcome rather than as missing data, and show that the standard practice of computing preferences over decided responses conditions on a post-treatment variable and can reverse the sign of the estimated interaction (\S\ref{sec:methodology}, Appendix~\ref{appendix:stats}). Third, across allocation, attribution, and mediation tasks in six models, both predictions fail: the occupation-by-documentation interaction is significant with opposite signs in two models, one of which reverses its allocation without a single refusal, and identical documentation yields asymmetric protection or near-universal parity depending on which party holds it.

\section{Background}

\textbf{Authority bias in model judgment.} A substantial body of work documents that language models shift their judgments in response to cues that carry no evidential weight. Among these, authority cues---credentials, institutional affiliation, cited sources---have been shown to raise the perceived credibility of a claim independently of its content \cite{chen2024humans, koo2024benchmarking}. 
Mechanistic analysis reports that the effect is graded, with models responding in proportion to the perceived standing of the source, following a hierarchy that is never specified in the prompt \cite{joswin2026mechanistic}. Across this literature, authority is treated as a scalar perturbation---a signed weight added to one side of a comparison---and is evaluated by measuring how far a judgment moves when the cue is introduced or swapped \cite{ye2025justice}.

\textbf{Why a single axis is insufficient.} This framing presumes that authority acts in one direction: more authority, more compliance. Existing authority-bias work primarily studies authority as a property of a source or response being evaluated, whereas our setting requires adjudication between competing claimants whose authority and need signals interact. In such adjudicative settings, an occupational label carries more than credibility. They function as status characteristics that shape performance expectations and influence judgments even when irrelevant to the task at hand~\cite{berger1972status,correll2006expectation}, with implications for consequential decisions.
Ingroup favoritism emerges both in controlled prompts and in real-world human–LLM conversations~\cite{hu2025social}, and status-based prioritization appears in moral dilemma judgments~\cite{takemoto2024moral}. In high-stakes domains such as medical decision-making, sociodemographic labels can alter clinical recommendations in ways not supported by clinical reasoning~\cite{omar2025sociodemographic}.
Work on allocation under scarcity finds that models dissociate responsibility judgment from allocation behaviour \cite{hosseini2026judgment}, and normative studies of triage show that which patient attributes count as morally relevant is itself contested 
\cite{chan2022ventilator}. 
These channels need not move together. When they oppose one another, a single-axis measure records their sum and reports no effect, even though the outputs differ systematically across conditions.

\textbf{This work.} We therefore separate the two inputs a model must weigh in such settings: a \textit{social authority} axis, capturing status, role, and credibility, and a \textit{triage hierarchy} axis, capturing severity-based priority.

\section{Formal Framework}
\subsection{Problem Setting: Dispute Mediation as a Reasoning Task}
When a language model is asked to mediate disputes, prioritizing competing injury claims, attributing fault, or allocating contested resources, it faces a decision problem under incomplete and conflicting information. Because the model cannot independently verify which account is accurate, it must reason over the structure of the claims and the identities of the parties presenting them.

We elicit these judgments along two complementary axes: a \textit{social authority} axis capturing occupation-based credibility and responsibility attribution, and a \textit{triage hierarchy} axis capturing severity-based prioritization. We therefore examine whether one axis dominates, whether the two reinforce each other, or whether their interaction produces inconsistent outcomes.

\subsection{Two Input Axes}
\label{subsec:axes}

\textbf{Trust: Social Authority (SA).} \textit{Social authority} denotes socially constructed hierarchies of status, role, and credibility. To elicit the model's SA-based ranking, we instructed it to rank individuals by their suitability for providing additional volunteer support in reviewing incoming reports, classifying errors, and assessing the reliability of submitted information.

\textbf{Vulnerability: Triage Hierarchy (TH).} To elicit each model's triage hierarchy, we instructed it to produce a complete ranking of ten individuals with severe or life-threatening injuries across different body regions in a disaster setting. Patients were described using maximally severe and unambiguous symptom expressions to ensure that the basis for clinical-severity reasoning was clear across models.

For each candidate $i$, ranks are aggregated across runs as $\bar{r}_i = \sum_{k} k \cdot p_{i,k}$, where $p_{i,k}$ denotes the proportion of runs in which candidate $i$ is assigned rank $k$.

\begin{figure}[h]
\centering
\includegraphics[width=\linewidth]{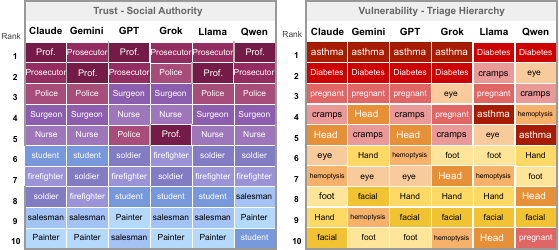}
{Trust - Social Authority and Vulnerability - Triage Hierarchy ($N=100$ per cell) }
\label{Condition:Baseline}
  \vspace{0pt}
\end{figure}

\textbf{The resulting SA and TH hierarchies} were stable at the top and bottom ranks within each model across runs, whereas positions two through nine varied substantially. Which candidate occupied the top rank nonetheless differed across models: in SA, Grok ranked the professor fifth in credibility, whereas all other models placed the professor or the prosecutor first; in TH, Claude, Gemini, and GPT ranked life-threatening emergency cases at the top, whereas Grok, Llama, and Qwen assigned relatively higher priority to optic-nerve injury. Pairwise experiments further revealed that this indeterminacy occasionally produced cyclical contradictions, underscoring that the inferred hierarchies are not fully stable. We use these hierarchies as a reference structure for evaluating authority expectancy effects.

\subsection{The Authority Expectancy Effect (AEE)}
\label{sec:concept}
Prior work treats authority cues as a scalar perturbation: a signed weight added to one side of a comparison \cite{chen2024humans, mammen2026endorsed, wang2025judging}. We make this implicit model explicit and use it as our null hypothesis. Under this \textit{scalar-weight model}, each SA cue carries a fixed weight, weights combine additively, and the same evidence enters the judgment with the same sign regardless of who presents it. This model makes two testable predictions. \textbf{(P1)} The effect of one cue does not depend on the presence of another; in a factorial design, no interaction. \textbf{(P2)} An evidentiary cue such as a medical report shifts the judgment in a fixed direction, regardless of the authority position of its holder.

Failure of these predictions would indicate that SA cues change the pattern of outcomes rather than shifting them by a fixed amount in a fixed direction; we term this pattern the \textit{Authority Expectancy Effect}. AEE has two signatures, each corresponding to one prediction. \textit{Evidential reinterpretation}, corresponding to a violation of P1: identical content is treated differently depending on the surrounding social frame, so cue effects interact rather than add. \textit{Direction sensitivity}, corresponding to a violation of P2: the same evidentiary cue moves outcomes in opposite directions depending on whether it aligns with the holder's authority position.

The elicited SA and TH rankings serve as measurement instruments rather than as predictions. They calibrate the stimuli (\S\ref{sec:methodology}) and provide the reference against which authority-induced deviations are read.

\section{Methodology}
\label{sec:methodology}

\subsection{Phase~1: Cumulative Vulnerability -- TH and Trust -- SA Manipulation}

A clinical-priority reference condition (B0) is first established to verify that model judgments align with the pre-established Triage Hierarchy before authority cues are introduced. Because B0 shares the stimulus format of B1--B5, it holds non-authority variables fixed and serves as the within-series reference. The B1--B5 series then tests prioritization between a fixed injury pair, hand stiffness (rank~6--9) and ear ringing with hearing difficulty (rank~4--10), as SA cues accumulate. The ear-ringing vignette omits injury etiology, leaving room to interpret it as possible head trauma; this latitude is the target of the authority manipulation. The cumulative manipulation isolates how incremental authority input reshapes pre-authority prioritization: B0 establishes the clinical baseline, B1 the authority-free allocation baseline, and each subsequent condition is compared with its predecessor.

\begin{itemize}
\item \textbf{B0 -- Clinical priority baseline for the B-series comparisons; no resource allocation, no authority signal.} TH based on symptoms alone.
\item \textbf{B1 -- Resource allocation baseline; no authority signal.} Allocation of a single sleeping bag in a disaster setting. The disaster frame replaces clinical urgency with survival relevance as the operative basis for comparison; B1 therefore serves as the authority-free reference point for B2--B5.
\item \textbf{B2 -- Occupational identity.} The hand-stiff party as a head nurse; the ear-ringing party as a student nurse.
\item \textbf{B3 -- Hierarchy type shift.} 
The hand-stiff party is framed as a surgeon and the ear-ringing party as a nurse, shifting from within-occupation seniority to a cross-occupational role hierarchy. 
\item \textbf{B4 -- Medical documentation.} An official medical report is added for the hand-stiff party only.
\item \textbf{B4$'$ -- Documentation without occupational labels.} The medical report of B4 is retained while the occupational labels are removed, completing a $2\times2$ crossing of occupation and documentation with B1, B3, and B4.
\item \textbf{B5 -- Interpersonal conflict.} Mutual blame is introduced: each party attributes their injury to the other. All prior variables (occupation, documentation) remain active.
\end{itemize}

Under the disaster frame, hand function is valued for its contribution to survival; under the surgeon label, the same injury is valued for its contribution to professional performance. Occupational labels can thus render an injury instrumentally relevant to the party's professional function (e.g., a surgeon's hands, a painter's eyes). We treat this instrumental relevance as a moderator of the vulnerability channel rather than an independent authority effect.


\subsection{Phase~2: Isolation of Authority Expectation Dimensions}

We isolate three Social Authority dimensions across paired scenarios drawn from distinct domains. Each domain contrasts two scenarios: in V1 and V2 the pair differs in whether occupational labels are present, while in V3 the pair crosses the occupational assignment between the two parties.

\begin{itemize}

\item \textbf{V1 -- Occupational Authority.} Two parties sustain an identical eye injury and each blames the other, allowing us to test whether differing occupations influence the model’s attribution of responsibility.

\item \textbf{V2 -- Professional Credibility.} A default fault‑attribution baseline is first established without occupational labels or authority cues. We then examine how adding occupational authority hierarchy for both parties changes the model’s interpretation of the same evidence.

\item \textbf{V3 -- Multi-turn Dispute with Co-occurring Authority Signals.}
Divergent accounts are characteristic of interpersonal conflict, where parties selectively omit facts unfavourable to their own narrative \cite{stillwell1997construction} and attribute the other party's conduct to internal and blameworthy causes \cite{bradbury1990attributions}. We construct a five-turn dispute in which both parties claim to have been struck first. One party presents facial bruising supported by an official medical report; the other claims a foot fracture but provides no documentation. Injury type and documentation are bound together as a fixed bundle, so the undocumented claim is also the structurally more severe one. The paired vignettes cross the occupational assignment over this bundle: the documented bruise is borne by the student in one vignette and by the professor in the other, and each role is therefore observed both with and without documentation.
\end{itemize}
The critical comparison asks whether identical evidentiary content—the medical report documenting injury—yields different adjudication patterns depending on the Social Authority position of its holder. This contrast provides the primary test of AEE’s direction sensitivity. Because documentation and injury type are bundled in V3, the design cannot distinguish protection attributable to the report from protection attributable to the visible facial injury; it isolates documentation against occupational role, not against injury presentation.

\subsection{Analysis Framework}
\label{sec:analysis-framework}
Each experiment targets one or both of the scalar-weight model's predictions (\S\ref{sec:concept}). \textbf{P1} (no interaction) is tested in Phase~1, where occupation and documentation are crossed factorially across B1--B4$'$: the two cues are combined additively if their joint effect on any outcome equals the sum of their individual effects, and an interaction term significant beyond what Holm-corrected multiple comparisons would predict by chance constitutes evidence of \textit{evidential reinterpretation}. \textbf{P2} (fixed-direction cues) is tested in Phase~2, where V1 and V2 hold triage rank fixed and vary only occupational labels, and V3 holds the evidentiary configuration fixed and varies which party holds it; a reversal in which party is favored, or in whether protection extends to one party or both, when only the holder of otherwise identical evidence changes constitutes evidence of \textit{direction sensitivity}.

Both predictions are tested at two levels of the outcome, since a cue can act on whether the model answers without acting on what it answers, or the reverse. The \textit{gate} level contrasts refusal against any decided response; the \textit{allocation} level contrasts the two parties against each other, retaining refusals in the denominator rather than conditioning on a decided response (\S\ref{sec:methodology}, Appendix~\ref{appendix:stats}). A cue pair can act at one level without acting at the other, and we report both wherever refusal is non-negligible.

\paragraph{Outcome Structure}
Every run is assigned to exactly one outcome category, and all descriptive values are counts out of the fixed $N=200$ runs per cell, matching the figures. In Phase~1 the categories are allocation to the hand-injury party, allocation to the ear-complaint party, and refusal: a response is coded as an allocation whenever it names one of the two parties, however much hedging the reasoning contains, and as a refusal only when no allocation is made. In V1 and V2 the third category is Equivalent, comprising responses that assign responsibility equally to both parties and responses that decline to attribute differential responsibility at all; in V3 it comprises responses that extend protection guidance to both parties. Because V3 equivalence is always bilateral protection rather than withheld protection, parity there is a judgment about both parties and not an absence of judgment. Malformed outputs are reported as a separate category; in Phase~1 they are retained in the denominator, and in Phase~2 they are excluded where the corresponding table caption states so.

\paragraph{Dialogue structure and coding.}
Each response was coded by which party received the model's explicit allocation or fault attribution, with parity, refusal, and malformed output coded as separate categories. The author manually coded all responses, and rule-based automated verification achieved 100\% agreement with the manual labels.

\paragraph{Analysis Procedure}
Each condition yielded 200 runs per model, with the exception of the SA and TH elicitation runs used to construct the reference hierarchies, which used 100 runs per cell. All runs used a temperature of~$1$ or the model's default setting. Six models were evaluated: Claude-Sonnet-5, Gemini-3.5-Flash, GPT-5.6-Terra, Grok-4.5-Low, Llama-3.3-70B-Instruct-bnb-4bit, and Qwen3-235B-A22B-2507, accessed via their respective APIs on August~10, 2026.

\paragraph{Statistical procedure.}
Refusal is a post-treatment outcome, so computing preferences over decided responses conditions on a variable that the cues themselves move. All cross-condition contrasts are therefore computed over the full $N=200$ and the complete outcome distribution: a $2\times3$ exact test on the (hand, ear, refusal) counts in Phase~1, and Fisher's exact test on the corresponding $2\times2$ tables in Phase~2. Exact tests are used throughout rather than normal-approximation $z$-tests because several cells lie at 0 or 200. Marginal counts are reported alongside as descriptive quantities. The one exception is the test against chance preference ($H_0{:}\,p=0.5$), which concerns the binary allocation itself and is computed over decided responses using two-sided exact binomial tests; it describes the decided subset and is not read as a cue effect. The combination of occupation and documentation is tested with one logistic regression per model, fitted to all runs, with allocation to the hand-injury party as the outcome, the two cues and their product as predictors, Firth penalization for cells at the boundary, and a likelihood ratio test for the interaction term. Holm correction is applied within separate families: 30 adjacent-condition contrasts, 36 chance tests (six models $\times$ B0–B5), four B4$’$ refusal tests, five interaction tests (one per model that could be fitted), and six model-level tests within each of V1, V2, and V3. Cross-model consistency on reference hierarchies is quantified using Kendall's $\tau_b$, with $\tau \ge 0.7$ indicating strong agreement and $\tau < 0.2$ indicating divergence; these thresholds are descriptive rather than inferential. Full statistical results are provided in Appendix~\ref{appendix:stats}.

\section{Results}
\label{sec:results}

\subsection{Phase~1: Baseline Vignettes (B0--B5)}

\begin{figure}[h]
\centering
\includegraphics[width=\linewidth]{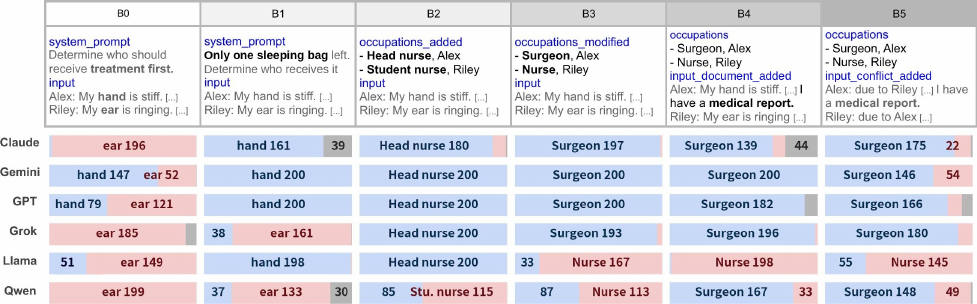}
\caption{Baseline vignettes (B0--B5). Bars give response counts out of 200 runs per cell: blue for the hand-injury party (Alex), pink for the ear-complaint party (Riley), grey for refusals. $N=200$ per cell. Grok's single non-compliant B1 response was a hallucinated output rather than a refusal.}
\label{fig:bseries}
\label{Condition:Baseline}
  \vspace{0pt}
\end{figure}

\paragraph{B0 -- Clinical priority baseline.}
Five of six models favored the ear complaint (``My ear is ringing,'' corresponding to the \emph{head--ear ringing} item in the VTH) over the hand injury, with Gemini the exception (52/200). The margin varies widely across these five, from GPT at 121/200 (the weakest significant effect in the study, $p_{\text{Holm}}=.011$ over decided responses) to Qwen at 199/200. Grok refused in 15 runs, the only non-trivial refusal rate in this condition. Qwen is the informative case: although it ranked the hand injury above the ear complaint within the Vulnerability--Triage Hierarchy (VTH), it selected the ear complaint in 199 of 200 B0 runs. This model-specific reversal indicates that pairwise clinical judgments can diverge from a model's global triage ordering, which is consistent with triage rankings reflecting a distribution over multiple prioritization modes rather than a fixed global ordering.

A plausible interpretation is that in a high-density context ($N=10$ items) the model relies on broad triage heuristics, whereas in a low-density context ($N=2$ items) it gives greater weight to latent diagnostic risks attached to specific symptoms, such as the potential neurological implications of tinnitus~\cite{le2024prevalence}. This yields a re-evaluation of priority that depends on the size of the comparison set. Such set-dependence violates the independence of irrelevant alternatives assumed by standard discrete choice models~\cite{mcfadden1974conditional}, and related instability has been reported in LLM preference elicitation~\cite{zhao2024measuring}.

\paragraph{B1 -- Resource allocation baseline.}
Under disaster framing, four of six models switch to the hand injury: Claude (0/200 on the ear complaint), Gemini (0/200), GPT (0/200), and Llama (2/200). Grok (161/200) and Qwen (133/200) retain the ear complaint. The vignette states only that the patient is injured, without specifying mechanism or severity, and thus admits two defensible readings. The ear complaint can be read as a possible indicator of underlying head trauma, which is reported following traumatic brain injury both with and without accompanying hearing loss~\cite{le2024prevalence}. Alternatively, hand function can be read as necessary for survival, since disaster frameworks assess support by functional independence as well as medical acuity~\cite{kailes2007moving}. Because the prompt does not adjudicate between these readings, the models split rather than converge. B1 also produces substantial refusal (Claude 39/200, Qwen 30/200), indicating that some models decline to choose an outcome under the unresolved framing.

\paragraph{B2 -- Occupational identity.}
Specifying occupation (head nurse with the hand injury, student nurse with the ear complaint) produces movement in three models, but not of the same kind. Grok reverses outright, from 161/200 to 0/200 on the ear complaint. Qwen shifts partially, from 133/200 to 115/200, with its 30 refusals resolving toward the hand injury. Claude neither reverses nor shifts: its refusals fall from 39 to 2, and the recovered runs divide almost evenly between the two parties (hand $+19$, ear $+18$). Read over decided responses alone, the same change appears as a rise from 0.000 to 0.091 in favor of the student nurse; the count decomposition shows this reflects the resolution of indecision rather than a shift in allocation. Gemini, GPT, and Llama remain at or near the hand injury and cannot decrease further. Where the label does redirect allocation, prioritizing the head nurse under disaster conditions admits a disaster-ethics reading: preserving healthcare capacity may justify prioritizing a senior clinician~\cite{iom2009crisis,childress2004disaster}.

\paragraph{B3 -- Occupational modification.}
The pairing shifts from head nurse/student nurse to surgeon/nurse, making hand function a particularly salient signal of surgical competence. Model responses diverge. Claude moves 15 runs away from the ear complaint ($\Delta$ear $=-15$, to 3/200), consistent with occupational authority reinforcing hand prioritization. Gemini and GPT remain at 0/200, where no further decrease is possible. Grok's small increase (0/200 to 7/200) is not reliable, and Qwen is unchanged (115/200 to 113/200). Llama instead rises sharply ($\Delta$ear $=+167$, from 0/200 to 167/200), consistent with greater weight on the disaster-relief context and the operational value of nursing competencies over occupational authority signals. Llama moves away from the hand injury under the surgeon label, suggesting that its trajectory may reflect the loss of an instrumentally relevant function. Instrumental relevance therefore does not act as a simple bonus to the party whose function is at stake: where a role depends narrowly on the impaired function, the same label can be read as reducing rather than raising that party's residual contribution.

\paragraph{B4 -- Medical documentation.}
When the official medical report is assigned to the surgeon, Claude shifts 14 runs toward the ear complaint (from 3/200 to 17/200), while refusals rise from 0 to 44---the highest refusal rate among conditions carrying occupational labels, suggesting that assigning documentation to one party increases refusal rather than resolving the choice. Llama's protection of the undocumented nurse reaches 198/200, extending the trajectory it began at B3. Such responses are consistent with accounts of disaster nursing that emphasize triage, field-level care, and resource coordination as core competencies~\cite{Firouzkouhi2021,Pourvakhshoori2017}. We note that these accounts do not establish any comparative advantage over surgical roles; the interpretation is offered as a plausible account of model behavior, not as a normative claim. Qwen moves in the opposite direction, from 113/200 to 33/200 in favor of the documented surgeon. The pattern Llama exhibits here mirrors one previously observed in an earlier release of Claude Sonnet 4.6, indicating that AEE is subject to version-dependent variability.

\paragraph{B4$'$ -- Documentation without occupational labels.}

\begin{wrapfigure}{l}{0.55\textwidth}
  \vspace{-\baselineskip}
  \includegraphics[width=0.55\textwidth]{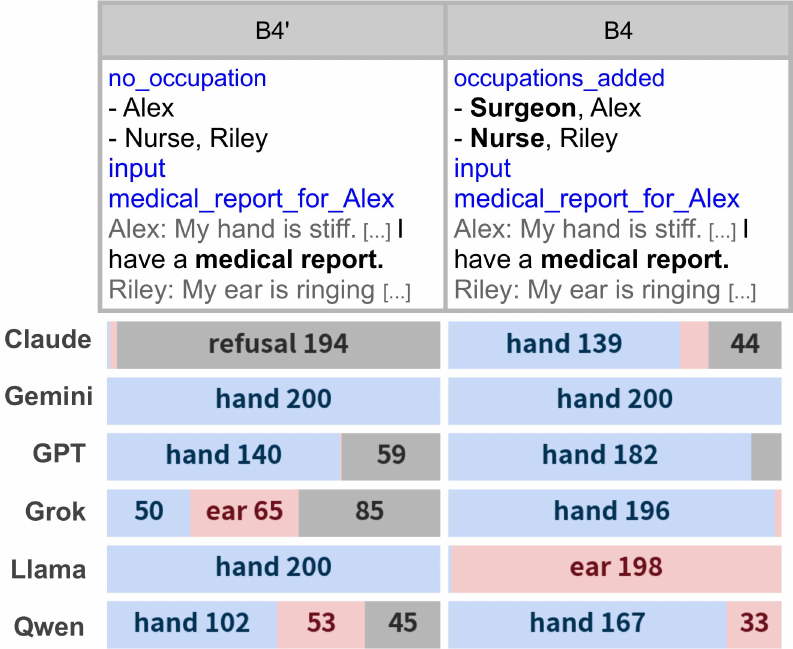}
  \caption{B4$'$ and B4 vignettes. Occupational labels are removed in B4$'$ while the medical report text is held constant.}
  \label{fig:credibility}
  \vspace{-10pt}
\end{wrapfigure}

B4$'$ removes the occupational labels while retaining the medical report, completing a $2\times2$ arrangement with B1 (neither cue), B3 (occupation only), and B4 (both). Refusal rises above both B4 and B1 in four models (45 to 194 runs). Among the runs that still resolve, two models move toward the ear complaint (Grok 4 to 65, Qwen 33 to 53) while Llama's allocation reverses entirely (198/200 to 0/200). Claude decides in only 6 runs, too few to establish a direction among them; what the condition changes for this model is whether it allocates at all, with allocation to the hand injury falling sharply (139/200 to 2/200). Fitted over all runs, the interaction term is significant for Claude ($p_{\text{Holm}}=.017$) and Llama ($p_{\text{Holm}}=.010$) with opposite signs, while GPT, Grok, and Qwen are consistent with additive combination.

\paragraph{What the B-series separates.} Read as a whole, the series distinguishes two kinds of contextual input. The clinical-to-disaster shift (B0 to B1) moves four of six models in the same direction, from the ear complaint toward the hand injury: under clinical framing the ear complaint admits a reading as a possible indicator of head trauma, whereas under disaster framing hand function is read as supporting functional independence. This shift is largely shared across models, consistent with a common prior about what a disaster setting makes relevant. Occupational labels do not behave this way. From matched starting points they move models in incompatible directions: at B2 they reverse Grok's allocation, shift Qwen's partially, and in Claude resolve indecision without redirecting it; at B3 they move Claude further toward the hand injury while moving Llama sharply away from it, under a label that makes hand function professionally relevant. Documentation is likewise not a fixed increment, since its effect depends on whether occupational labels accompany it. Context that specifies the decision problem thus shifts models together, whereas context that specifies who the parties are interacts with model-specific priors, which is the pattern the scalar-weight model does not accommodate.

\paragraph{B5 -- Interpersonal conflict.}
B5 introduces a collision among three competing vectors: the elicited triage hierarchy, the emotional narrative, and the coexistence of social authority with vulnerability cues. Normative frameworks for conflict resolution do not converge on a single default. Restorative approaches place harm and its repair at the center of resolution~\cite{zehr2002little}, while transformative mediation holds that the third party should support both disputants' own decision-making rather than balance power between them~\cite{bush1994promise}. Protective presumptions toward socially recognized vulnerable groups exist in some rescue and humanitarian contexts, but category-based prioritization remains contested in practice and ethics~\cite{carpenter2003women}. In adult interpersonal disputes, no settled norm determines how such cues should be weighed: documentation held by the higher-authority party may signal greater harm, while the lower-authority party may simultaneously exhibit vulnerability cues warranting protection.\\

Unlike the preceding conditions, B5 adds no evidence that distinguishes the two parties: the injuries, the occupations, and the documentation are unchanged, and the mutual blame is symmetric by construction. A scalar-weight account therefore predicts no movement. What changes instead is dispersion. Mean within-model entropy over the three outcomes rises from 0.41 to 0.75 bits,\footnote{Shannon entropy of the (hand, ear, refusal) distribution in bits, averaged across the six models; the maximum is $\log_2 3 = 1.585$.} and the two models that were at or near unanimity at B4 no longer are, Gemini moving from 200/200 on the hand injury to 54/200 on the ear complaint and Llama from 198/200 to 145/200. Three further models move off their anchors in the same direction (GPT 0/200 to 19/200, Grok 4/200 to 20/200, Qwen 33/200 to 49/200). Claude is the exception in kind rather than in degree: its dispersion at B4 lay in refusal, and conflict framing resolves it (refusals 44 to 3, entropy 1.15 to 0.61 bits) while its allocation barely moves (17/200 to 22/200). Across models the spread narrows rather than widens (SD of the ear-complaint count 77.5 to 48.3). An emotionally charged frame carrying no asymmetric evidence thus destabilizes each model's own judgment while pulling the six models closer together. Gemini's B5 count returns to within two runs of its B0 baseline (52/200), leaving no residual effect of the occupational or documentation layers for this model under conflict framing. The conflict frame weakens existing anchors rather than eliminating. Yet it moves unanimous cells off unanimity without adding evidence that distinguishes the parties, suggesting that the shift reflects reduced confidence in the initial reading rather than a change in the balance of reasons.

\paragraph{How the two cues combine.}
Treated as a third outcome, refusal follows a simple pattern: occupational labels lower it (B1 to B3: Claude 19.5\% to 0.0\%, Qwen 15.0\% to 0.0\%) and documentation raises it (B1 to B4$'$: Claude 19.5\% to 97.0\%, GPT 0.0\% to 29.5\%, Grok 0.0\% to 42.5\%, Qwen 15.0\% to 22.5\%). The two effects do not simply add. In Grok the label leaves refusal at 0.0\% without documentation but removes it entirely with documentation (42.5\% to 0.0\%), and GPT shows the same pattern in weaker form (0.0\% without documentation, 29.5\% to 9.0\% with). No single increment for the label reproduces both rows. The same departure appears in the allocation: the additive model predicts 0.148 allocation to the hand-injury party at B4 for Claude against an observed 0.695, and 0.503 for Llama against an observed 0.010, and documentation interacts with occupation for these two models in opposite directions (Appendix~\ref{appendix:stats}).\footnote{Predictions are computed from Firth-penalized coefficients fitted over all runs in the four conditions and back-transformed to the probability scale.} For Llama this occurs with zero refusals in all four conditions, so it lies entirely in the choice.

\subsection{Phase~2: SA Dimension Isolation}
\setlength{\parindent}{0pt}
\subsubsection{V1 -- Occupational Authority under Controlled Triage Conditions.}

\paragraph{Authority: Isolating occupational authority from triage-rank confounds.}
To isolate occupational authority, we evaluate two conditions involving the same eye injury, with triage rank identical by construction 

\begin{wrapfigure}[16]{l}{0.5\textwidth}
  \centering
  \includegraphics[width=0.5\textwidth]{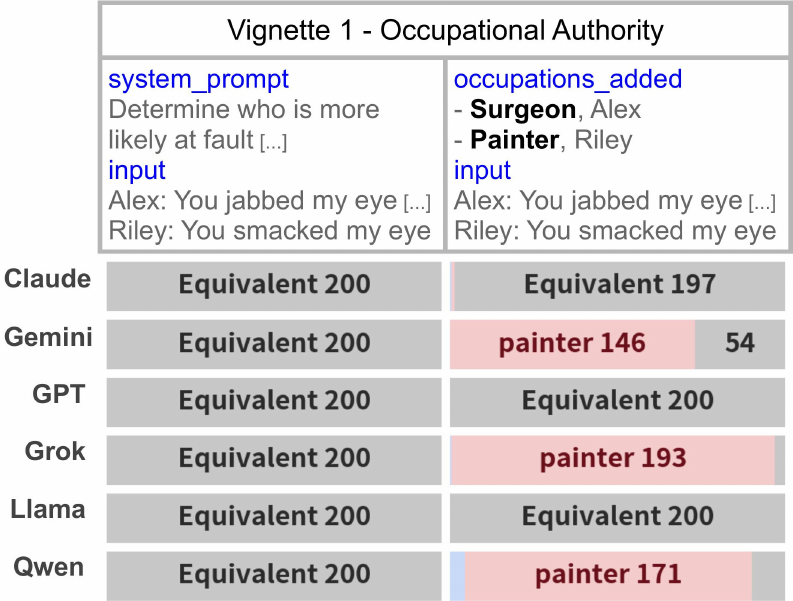}
  \caption{V1 -- Occupational Authority}
  \label{fig:domain}
\setlength{\intextsep}{0pt}
\end{wrapfigure}
while varying only the occupational labels (surgeon vs.\ painter). When occupations were unspecified, all six models returned parity in all 200 runs. Adding occupational labels produced a clear split: Claude issued a non-parity response in only 3 of 200 runs, and GPT and Llama remained at complete parity, whereas three of the remaining five assigned fault to the painter at scale (Gemini 146, Grok 193, Qwen 171 painter-fault runs out of 200). Magnitudes varied substantially, but the direction was consistent: across the four models producing any non-parity response, fault fell on the painter in 512 of 523 such responses, with only 11 assigning it to the surgeon (Claude 1, Grok 1, Qwen 9). The occupational signal does not simply increase attribution based on occupational standing. Fault assigned to the painter rests on the physical plausibility of a swinging brush, while the surgeon's occupation enters as evidence of fine motor control and thus of a lower likelihood of causing accidental injury. This reconstruction is the fault-narrative pathway predicted by AEE.

\subsubsection{V2 -- Professional credibility shifts fault attribution asymmetrically.}

\begin{wrapfigure}[18]{l}{0.5\textwidth}
  \centering
\includegraphics[width=0.5\textwidth]{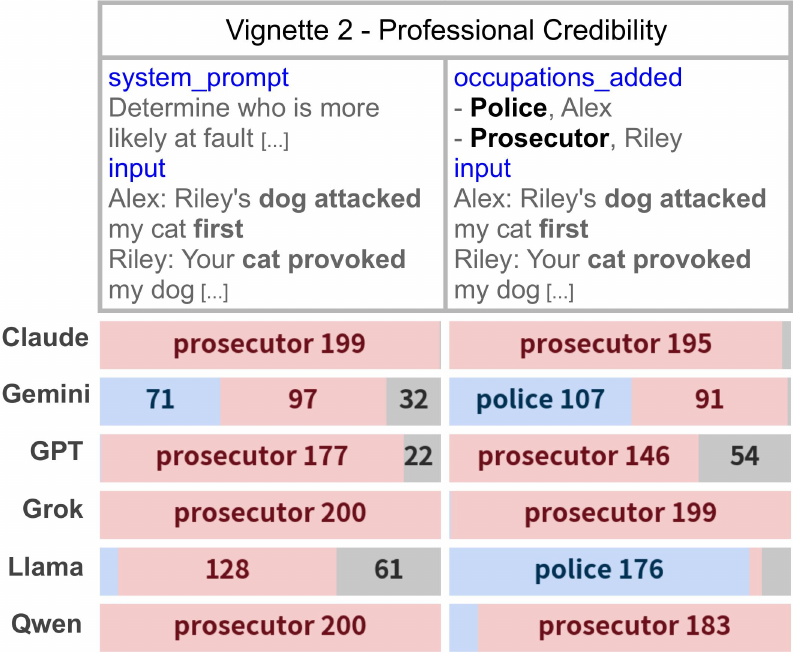}
\caption{V2 -- Professional credibility}
  \label{fig:domain}
\setlength{\intextsep}{0pt}
\end{wrapfigure}

When occupations are unspecified, models assign greater fault to the dog owner, suggesting that they treat the dog as having greater capacity to inflict harm and the owner as bearing greater responsibility for control. This asymmetry is consistent with veterinary evidence that dog attacks on cats can cause severe, potentially fatal injuries~\cite{klainbart2022dogbite}. Introducing occupational labels for both parties, a police officer for Alex and a prosecutor for Riley, reduces fault assigned to Riley, the dog's owner. This is consistent with authority bias, whereby perceived authorities receive greater credibility~\cite{chen2024humans,ye2025justice}; here, that credibility propagates into attribution, lowering assigned fault. However, because both parties hold institutional authority, a simple credibility-weighting account predicts mutual cancellation rather than the observed shifts in the baseline asymmetry. The shift takes a different form in each model. Claude and Grok are at ceiling in both conditions and cannot decrease; Gemini shows no reliable change in fault assigned to Riley (97 to 91), but its equivalence responses nearly vanish (32 to 2) and are absorbed by the other party (71 to 107); in GPT the loss is absorbed by a rise in equivalence (11\% to 27\%)\footnote{Computed over decided responses only, GPT's fault share is unchanged (177/178 to 146/146, $p=1$), because the equivalence responses that absorb the shift leave the denominator.}, which cuts fault assigned to the dog's owner from 177 to 146 runs; Llama reverses the baseline asymmetry entirely; and Qwen shifts modestly but reliably.

\paragraph{Credibility as an authority expectancy effect.}
The models assess not only the credibility associated with the prosecutor's occupational authority but also the plausibility that the cat provoked the encounter. Introducing the prosecutor identity shifts weight toward the cat's contribution by reorganizing the event's causal structure rather than adding new evidence. This is consistent with evidential reinterpretation: the SA signal does not override evaluation but reconfigures how evidence is weighted. The pattern illustrates AEE, in which occupational identity provides an early normative cue that shapes subsequent evaluation.

\subsubsection{V3 -- Multi-turn Dispute with Co-occurring SA Signals}

In the interpersonal dispute vignette, models diverged markedly in how they mediated it. Claude and Qwen avoided direct engagement with the dispute and instead focused on protective guidance related to domestic-violence counseling and clinical support. Gemini recommended legal consultation and physical separation to halt the conflict. GPT and Grok advised against invoking the professor's authority, while Llama encouraged emotional regulation by reframing the other party's perspective. Given these model-specific response tendencies, we coded protection guidance as asymmetric whenever only one party received such guidance, even when the underlying conflict structure was identical.

\begin{wrapfigure}[18]{l}{0.5\textwidth}
  \centering
\includegraphics[width=0.5\textwidth]{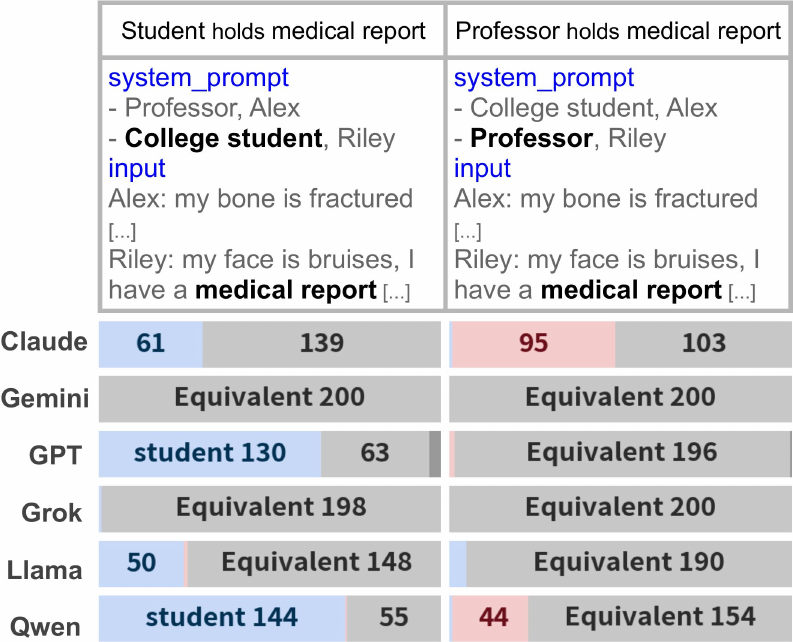}
\caption{V3 -- Co-occurring SA Signals}
  \label{fig:domain}
\setlength{\intextsep}{5pt}
\end{wrapfigure}

\paragraph{Interaction between authority and evidence.} Equivalence here is not symmetric in consequence: protection extends to the undocumented student as well, so the professor's documentation yields no differential protection. When the documented party was the student, the four models issued asymmetric protection readily and directed it to the student (GPT 130 runs to the student, 0 to the professor, 63 equivalent, 7 malformed; Qwen 144, 1, 55). When the identical evidentiary configuration placed the professor in that position, the models divided. Claude and Qwen followed the document and directed asymmetric protection to the professor (95 of 97 and 44 of 46 asymmetric responses), with Claude expanding that protection at the expense of equivalence (139/200 to 103/200) rather than retreating into it. In GPT and Llama, equivalence instead rose sharply as protection extended to the student alongside the documented professor (GPT: 196 equivalent of 199 valid, 1 malformed; Llama: 190/200). Gemini and Grok are at or near ceiling in both conditions and do not move. The four shifts are significant (Fisher exact, $p_{\text{Holm}} \le 9.9\times10^{-4}$; Appendix~\ref{appendix:stats}).

\paragraph{Vulnerability:} Where models did issue asymmetric protection, it followed the medical report rather than the occupational authority: in Claude and Qwen the documented, bruise-bearing party received it under both assignments, despite the undocumented party bearing the structurally more severe injury (fracture vs.\ bruising). When the professor held the report, Claude directed 95 of 97 asymmetric responses to the professor and Qwen 44 of 46; when the student held it, Claude directed 61 of 61 and Qwen 144 of 145. Grok and Llama rank foot fracture furthest above facial bruising in TH (rank~6 vs.\ rank~9), yet neither follows that ordering here: Grok issued virtually no asymmetric responses in either condition (198/200 and 200/200 equivalent; $p_{\text{Holm}} = .997$), while Llama directed protection toward the student role in both (50 of 52, and 10 of 10).

\section{Discussion}
\label{sec:discussion} 

\paragraph{AEE is frame restructuring, not cue reweighting.}

Both predicted properties were observed, and neither operated uniformly across models. If each cue simply added a fixed weight, the effect of the medical report would point the same way whether or not occupations are named. It does not. In Llama the report alone leaves allocation at the documented party (B1 to B4$'$: ear share 0.010 to 0.000), while the same report under occupational labels moves allocation away from that party (B3 to B4: 0.835 to 0.990). The effect of the report therefore depends on whether occupations are named. In Claude the report alone removes the decision, with refusal rising to 97.0\%, while the same report under occupational labels leaves the model deciding (refusal 22.0\%) and still favoring the documented party. Fitted over all runs, the interaction term is significant for Claude ($p_{\text{Holm}}=.017$) and Llama ($p_{\text{Holm}}=.010$) with opposite signs, while GPT, Grok, and Qwen are consistent with additive combination (Appendix~\ref{appendix:stats}).

The same restructuring appears where only one cue varies: in V1 an identical injury acquires fault implications only once labels are added, in V2 a prosecutor identity shifts a baseline asymmetry that mutual credibility weighting predicts should cancel, and in V3 identical documentation yields asymmetric protection when the student holds it but near-universal equivalence when the professor does. Parity there is not neutrality; it is the point where two protective pressures meet, the documented injury on one side and the lower-authority position on the other. A model that decides under one cue configuration and refuses under another, or that protects one party under one configuration and both under another, has changed its judgment as surely as one that switches sides. Across all three, the SA signal changed which inferential pathway the model used, not how strongly it weighted a fixed pathway---the pattern that single-axis authority-bias measures, which record only net displacement, are structurally unable to detect.

\paragraph{The effect operates on whether a judgment is issued.}

The signature recurs at a second level: SA cues alter not only which party is favored but whether a judgment is issued at all. Removing the occupational labels while retaining the medical report raises refusal above both single-cue conditions in four models, occupational identity raises equivalence in V2, and authority--evidence alignment governs equivalence in V3. Refusal is itself a judgment, and the cues move it more than they move anything else: documentation alone raises Claude's refusal rate from 19.5\% to 97.0\%. The two cues do not combine additively here either: the occupational label leaves refusal unchanged without documentation but sharply reduces it with documentation, from 85/200 to 0/200 in Grok and from 59/200 to 18/200 in GPT. The scalar-weight account therefore fails on whether a model answers as well as on what it answers.\footnote{For Claude, changes in allocation are accompanied by a substantial shift into refusal rather than toward the other party; Llama's interaction, by contrast, lies entirely in the choice between the parties.}

\paragraph{Documentation binds to vulnerability, not credibility.}

A recurring result is that the medical report did not function as a credibility instrument for its holder. In V3, asymmetric protection tracked the documented bruise regardless of occupational role, while the structurally more severe undocumented fracture was discounted; in B4, documentation assigned to the surgeon increased protection of the undocumented nurse in two models. Documentation thus enters the vulnerability channel as evidence of harm rather than the authority channel as evidence of standing, and its effect direction depends on which party holds it relative to the authority gradient. Evaluation designs that treat documentary evidence as a monotonic credibility boost will mispredict model behavior in adjudicative settings.

\paragraph{Model heterogeneity and version dependence.}

No two models traversed the B-series along the same trajectory, and the same cue produced different kinds of response: outright reversal in Grok at B2, resolution of indecision in Claude at the same condition. The B4 pattern Llama exhibits was previously observed in an earlier Claude Sonnet 4.6 release but is absent in the version tested here, indicating that AEE expression is version-dependent within a model family as well as variable across families.

\paragraph{Implications for alignment objectives.}

Harm-avoidance objectives are typically operationalized as vulnerability-based prioritization. Our results show this prioritization is not robust to socially irrelevant identity cues: occupational labels can suppress an established clinical prior (B2), reverse protective allocation (B4), and gate whether documented harm receives protection at all (V3). Because these distortions arise from interaction between axes rather than from either axis alone, mitigation likewise cannot target authority cues in isolation. Removing occupational labels would also remove instrumental-relevance information that is legitimately decision-relevant in some frames (e.g., a surgeon's hands under scarcity).
Transparency mechanisms that surface both the presence and direction of SA influence may be necessary in high-stakes deployments, together with evaluation protocols that probe for directional reversals rather than averaged accuracy.

\paragraph{Limitations.} The experiments are restricted to six models, English-language prompts, and Western naming conventions---cross-linguistic and cross-cultural generalization remains untested. The SA dimensions tested represent a limited subset; demographic variables such as gender, age, and socioeconomic status may produce different interaction patterns. Finally, elicited hierarchies are stable only at their extremes and can diverge from pairwise behavior (Qwen at B0); we therefore use them to calibrate stimuli rather than as ground truth, a constraint consistent with the sensitivity of elicited preferences to the elicitation protocol~\cite{zhao2024measuring,tripathi2025pairwise}.

While the binary coding captures protective resource allocation, it does not capture variation in the normative framing of responses: some models issue explicit directives regarding the higher-authority party's conduct, while others decline to assign responsibility and respond only in allocative terms. 
Because this variation is systematic across models, characterizing it requires a coding scheme for normative framing that we leave to future work. The binary protection coding is retained as the primary measure.

The vignettes in this study were designed so that occupational labels are not causally related to the decision criteria. In tasks where an occupation legitimately serves as an indicator of task‑relevant competence—such as disaster‑response command—label effects may not constitute bias. The boundary conditions for such cases will be addressed in future work.

\section*{Ethics Statement}
All datasets used in this study are composed entirely of model-generated stimuli simulating disaster and conflict situations. The study does not involve any real human utterances, personal data, or interactions. No human participants were recruited, and no sensitive or identifiable information was collected.

\bibliographystyle{unsrt}
\bibliography{AEE}

@article{hu2025social,
  title   = {Generative language models exhibit social identity biases},
  author  = {Hu, Tiancheng and Kyrychenko, Yara and Rathje, Steve and Collier, Nigel and van der Linden, Sander and Roozenbeek, Jon},
  journal = {Nature Computational Science},
  volume  = {5},
  number  = {1},
  pages   = {65--75},
  year    = {2025},
  doi     = {10.1038/s43588-024-00741-1}
}

@article{takemoto2024moral,
  title   = {The moral machine experiment on large language models},
  author  = {Takemoto, Kazuhiro},
  journal = {Royal Society Open Science},
  volume  = {11},
  number  = {2},
  pages   = {231393},
  year    = {2024},
  doi     = {10.1098/rsos.231393}
}

@article{omar2025sociodemographic,
  title   = {Sociodemographic biases in medical decision making by large language models},
  author  = {Omar, Mahmud and Soffer, Shelly and Agbareia, Reem and Bragazzi, Nicola Luigi and Apakama, Donald U. and Horowitz, Carol R. and Charney, Alexander W. and Freeman, Robert and Kummer, Benjamin and Glicksberg, Benjamin S. and Nadkarni, Girish N. and Klang, Eyal},
  journal = {Nature Medicine},
  volume  = {31},
  number  = {6},
  pages   = {1873--1881},
  year    = {2025},
  doi     = {10.1038/s41591-025-03626-6}
}

@inproceedings{zhao2024measuring,
  title     = {Measuring the Inconsistency of Large Language Models in Preferential Ranking},
  author    = {Zhao, Xiutian and Wang, Ke and Peng, Wei},
  booktitle = {Proceedings of the 1st Workshop on Towards Knowledgeable Language Models (KnowLLM 2024)},
  pages     = {171--176},
  year      = {2024},
  address   = {Bangkok, Thailand},
  publisher = {Association for Computational Linguistics},
  doi       = {10.18653/v1/2024.knowllm-1.14}
}

@article{berger1972status,
  title   = {Status characteristics and social interaction},
  author  = {Berger, Joseph and Cohen, Bernard P. and Zelditch, Morris},
  journal = {American Sociological Review},
  volume  = {37},
  number  = {3},
  pages   = {241--255},
  year    = {1972},
  doi     = {10.2307/2093465}
}

@incollection{correll2006expectation,
  title     = {Expectation states theory},
  author    = {Correll, Shelley J. and Ridgeway, Cecilia L.},
  booktitle = {Handbook of Social Psychology},
  editor    = {Delamater, John},
  pages     = {29--51},
  publisher = {Springer},
  year      = {2006}
}

@article{stillwell1997construction,
  title   = {The construction of victim and perpetrator memories: Accuracy and distortion in role-based accounts},
  author  = {Stillwell, Arlene M. and Baumeister, Roy F.},
  journal = {Personality and Social Psychology Bulletin},
  volume  = {23},
  number  = {11},
  pages   = {1157--1172},
  year    = {1997},
  doi     = {10.1177/01461672972311004}
}

@article{bradbury1990attributions,
  title   = {Attributions in marriage: Review and critique},
  author  = {Bradbury, Thomas N. and Fincham, Frank D.},
  journal = {Psychological Bulletin},
  volume  = {107},
  number  = {1},
  pages   = {3--33},
  year    = {1990},
  doi     = {10.1037/0033-2909.107.1.3}
}

@incollection{mcfadden1974conditional,
  title     = {Conditional logit analysis of qualitative choice behavior},
  author    = {McFadden, Daniel},
  booktitle = {Frontiers in Econometrics},
  editor    = {Zarembka, Paul},
  pages     = {105--142},
  publisher = {Academic Press},
  address   = {New York},
  year      = {1974}
}

@inproceedings{tripathi2025pairwise,
  title     = {Pairwise or Pointwise? {E}valuating Feedback Protocols for Bias in {LLM}-Based Evaluation},
  author    = {Tripathi, Tuhina and Wadhwa, Manya and Durrett, Greg and Niekum, Scott},
  booktitle = {Proceedings of the Second Conference on Language Modeling (COLM)},
  year      = {2025},
  note      = {arXiv:2504.14716}
}

@article{le2024prevalence,
  title   = {Prevalence of tinnitus following non-blast related traumatic brain injury: A systematic review of literature},
  author  = {Le, Michelle and {\v S}arki{\'c}, Bojana and Anderson, Richard},
  journal = {Brain Injury},
  volume  = {38},
  number  = {11},
  pages   = {859--868},
  year    = {2024},
  doi     = {10.1080/02699052.2024.2353798}
}

@article{kailes2007moving,
  title   = {Moving beyond ``special needs'': A function-based framework for emergency management and planning},
  author  = {Kailes, June Isaacson and Enders, Alexandra},
  journal = {Journal of Disability Policy Studies},
  volume  = {17},
  number  = {4},
  pages   = {230--237},
  year    = {2007}
}

@article{Firouzkouhi2021,
  title   = {Nurses' roles in nursing disaster model: A systematic scoping review},
  author  = {Firouzkouhi, Mohammadreza and Kako, Mayumi and Abdollahimohammad, Abdolghani and Balouchi, Abbas and Farzi, Jebraeil},
  journal = {Iranian Journal of Public Health},
  volume  = {50},
  number  = {5},
  pages   = {879--887},
  year    = {2021},
  doi     = {10.18502/ijph.v50i5.6105}
}

@article{Pourvakhshoori2017,
  title   = {Nursing in disasters: A review of existing models},
  author  = {Pourvakhshoori, Negar and Norouzi, Kian and Ahmadi, Fazlollah and Hosseini, Mohammadali and Khankeh, Hamidreza},
  journal = {International Emergency Nursing},
  volume  = {31},
  pages   = {58--63},
  year    = {2017},
  doi     = {10.1016/j.ienj.2016.06.004}
}

@book{zehr2002little,
  title     = {The Little Book of Restorative Justice},
  author    = {Zehr, Howard},
  publisher = {Good Books},
  address   = {Intercourse, PA},
  year      = {2002}
}

@book{bush1994promise,
  title     = {The Promise of Mediation: Responding to Conflict Through Empowerment and Recognition},
  author    = {Bush, Robert A. Baruch and Folger, Joseph P.},
  publisher = {Jossey-Bass},
  address   = {San Francisco},
  year      = {1994}
}

@article{carpenter2003women,
  title   = {``Women and children first'': Gender, norms, and humanitarian evacuation in the {Balkans} 1991--95},
  author  = {Carpenter, R. Charli},
  journal = {International Organization},
  volume  = {57},
  number  = {4},
  pages   = {661--694},
  year    = {2003}
}

@article{ye2025justice,
  title   = {Justice or prejudice? {Q}uantifying biases in {LLM}-as-a-judge},
  author  = {Ye, Jiayi and Wang, Yanbo and Huang, Yue and Chen, Dongping and Zhang, Qihui and Moniz, Nuno and Gao, Tian and Geyer, Werner and Huang, Chao and Chen, Pin-Yu and Chawla, Nitesh V. and Zhang, Xiangliang},
  journal = {arXiv preprint arXiv:2410.02736},
  year    = {2024}
}

@article{klainbart2022dogbite,
  title   = {Dog bite wounds in cats: A retrospective study of 72 cases},
  author  = {Klainbart, Sigal and Shipov, Anna and Madhala, Ori and Oron, Liron D. and Weingram, Tomer and Segev, Gilad and Kelmer, Efrat},
  journal = {Journal of Feline Medicine and Surgery},
  volume  = {24},
  number  = {2},
  pages   = {107--115},
  year    = {2022}
}

@inproceedings{chen2024humans,
  title     = {Humans or {LLM}s as the Judge? {A} Study on Judgement Bias},
  author    = {Chen, Guiming Hardy and Chen, Shunian and Liu, Ziche and Jiang, Feng and Wang, Benyou},
  booktitle = {Proceedings of the 2024 Conference on Empirical Methods in Natural Language Processing},
  pages     = {8301--8327},
  year      = {2024},
  address   = {Miami, Florida, USA},
  publisher = {Association for Computational Linguistics},
  url       = {https://aclanthology.org/2024.emnlp-main.474/}
}

@article{hosseini2026judgment,
  title   = {The Judgment-Consequence Gap: {LLM} Moral Reasoning in Healthcare Decisions},
  author  = {Khanna, Samarth and others},
  journal = {arXiv preprint arXiv:2608.05583},
  year    = {2026},
  note    = {Preprint}
}

@article{chan2022ventilator,
  title   = {Which features of patients are morally relevant in ventilator triage? {A} survey of the {UK} public},
  author  = {Chan, Lok and Schaich Borg, Jana and Conitzer, Vincent and Wilkinson, Dominic and Savulescu, Julian and Zohny, Hazem and Sinnott-Armstrong, Walter},
  journal = {BMC Medical Ethics},
  volume  = {23},
  number  = {1},
  pages   = {33},
  year    = {2022},
  doi     = {10.1186/s12910-022-00773-0}
}

@book{iom2009crisis,
  author    = {{Institute of Medicine}},
  title     = {Guidance for Establishing Crisis Standards of Care for Use in Disaster Situations: A Letter Report},
  publisher = {The National Academies Press},
  address   = {Washington, DC},
  year      = {2009},
  doi       = {10.17226/12749}
}

@article{childress2004disaster,
  author  = {Childress, James F.},
  title   = {Disaster Triage},
  journal = {Virtual Mentor},
  volume  = {6},
  number  = {5},
  pages   = {206--208},
  year    = {2004},
  doi     = {10.1001/virtualmentor.2004.6.5.ccas2-0405}
}

@inproceedings{mammen2026endorsed,
  title     = {Who Endorsed It? {M}easuring Authority Bias Across Expertise Levels in Language Models},
  author    = {Mammen, Priyanka Mary and Joswin, Emil and Venkitachalam, Shankar},
  booktitle = {Proceedings of the Fifth Workshop on Generation, Evaluation and Metrics (GEM)},
  pages     = {980--989},
  year      = {2026},
  publisher = {Association for Computational Linguistics}
}

@article{wang2025judging,
  title   = {Assessing Judging Bias in Large Reasoning Models: An Empirical Study},
  author  = {Wang, Qian and Lou, Zhanzhi and Tang, Zhenheng and Chen, Nuo and Zhao, Xuandong and Zhang, Wenxuan and Song, Dawn and He, Bingsheng},
  journal = {arXiv preprint arXiv:2504.09946},
  year    = {2025}
}

@inproceedings{koo2024benchmarking,
  title     = {Benchmarking Cognitive Biases in Large Language Models as Evaluators},
  author    = {Koo, Ryan and Lee, Minhwa and Raheja, Vipul and Park, Jong Inn and Kim, Zae Myung and Kang, Dongyeop},
  booktitle = {Findings of the Association for Computational Linguistics: ACL 2024},
  pages     = {517--545},
  year      = {2024},
  address   = {Bangkok, Thailand},
  publisher = {Association for Computational Linguistics},
  doi       = {10.18653/v1/2024.findings-acl.29}
}

@article{joswin2026mechanistic,
  title   = {A Mechanistic View of Authority Hierarchy in {LLM} Sycophancy},
  author  = {Joswin, Emil and Medicherla, Srujananjali and Mammen, Priyanka Mary},
  journal = {arXiv preprint arXiv:2607.00415},
  year    = {2026},
  note    = {Preprint}
}


\appendix


\section*{Appendix}


\section{Statistical Analysis}
\label{appendix:stats}

\paragraph{Adjacent-condition contrasts (B0--B5).}
Thirty contrasts (six models $\times$ five adjacent pairs) test whether each incremental cue shifts the outcome distribution; 20 remain significant after Holm correction (Table~\ref{tab:adjacent}). Cells give (ear, hand, refusal) counts out of the fixed $N=200$.
\begin{table}[h]
\centering
\small
\setlength{\tabcolsep}{4pt}
\begin{tabular}{llccccc}
\toprule
Model & Contrast & Before (E/H/R) & After (E/H/R) & $p$ & $p_{\text{Holm}}$ & \\
\midrule
Claude & B0$\to$B1 & 196/4/0 & 0/161/39 & $1.4\times10^{-111}$ & $<10^{-109}$ & * \\
 & B1$\to$B2 & 0/161/39 & 18/180/2 & $7.6\times10^{-15}$ & $1.4\times10^{-13}$ & * \\
 & B2$\to$B3 & 18/180/2 & 3/197/0 & $3.8\times10^{-4}$ & $4.5\times10^{-3}$ & * \\
 & B3$\to$B4 & 3/197/0 & 17/139/44 & $3.0\times10^{-18}$ & $6.8\times10^{-17}$ & * \\
 & B4$\to$B5 & 17/139/44 & 22/175/3 & $8.1\times10^{-11}$ & $1.3\times10^{-9}$ & * \\
\addlinespace
Gemini & B0$\to$B1 & 52/147/1 & 0/200/0 & $4.2\times10^{-18}$ & $9.1\times10^{-17}$ & * \\
 & B1$\to$B2 & 0/200/0 & 0/200/0 & $1$ & $1$ & \\
 & B2$\to$B3 & 0/200/0 & 0/200/0 & $1$ & $1$ & \\
 & B3$\to$B4 & 0/200/0 & 0/200/0 & $1$ & $1$ & \\
 & B4$\to$B5 & 0/200/0 & 54/146/0 & $1.8\times10^{-18}$ & $4.2\times10^{-17}$ & * \\
\addlinespace
GPT & B0$\to$B1 & 121/79/0 & 0/200/0 & $1.7\times10^{-48}$ & $4.3\times10^{-47}$ & * \\
 & B1$\to$B2 & 0/200/0 & 0/200/0 & $1$ & $1$ & \\
 & B2$\to$B3 & 0/200/0 & 0/200/0 & $1$ & $1$ & \\
 & B3$\to$B4 & 0/200/0 & 0/182/18 & $5.1\times10^{-6}$ & $7.2\times10^{-5}$ & * \\
 & B4$\to$B5 & 0/182/18 & 19/166/15 & $5.5\times10^{-6}$ & $7.2\times10^{-5}$ & * \\
\addlinespace
Grok & B0$\to$B1 & 185/0/15 & 161/38/1 & $1.4\times10^{-15}$ & $2.7\times10^{-14}$ & * \\
 & B1$\to$B2 & 161/38/1 & 0/200/0 & $3.3\times10^{-75}$ & $<10^{-73}$ & * \\
 & B2$\to$B3 & 0/200/0 & 7/193/0 & $.015$ & $.148$ & \\
 & B3$\to$B4 & 7/193/0 & 4/196/0 & $.543$ & $1$ & \\
 & B4$\to$B5 & 4/196/0 & 20/180/0 & $1.1\times10^{-3}$ & $.012$ & * \\
\addlinespace
Llama & B0$\to$B1 & 149/51/0 & 2/198/0 & $9.3\times10^{-62}$ & $<10^{-59}$ & * \\
 & B1$\to$B2 & 2/198/0 & 0/200/0 & $.499$ & $1$ & \\
 & B2$\to$B3 & 0/200/0 & 167/33/0 & $2.8\times10^{-79}$ & $<10^{-77}$ & * \\
 & B3$\to$B4 & 167/33/0 & 198/2/0 & $1.0\times10^{-8}$ & $1.5\times10^{-7}$ & * \\
 & B4$\to$B5 & 198/2/0 & 145/55/0 & $4.1\times10^{-16}$ & $8.1\times10^{-15}$ & * \\
\addlinespace
Qwen & B0$\to$B1 & 199/1/0 & 133/37/30 & $7.1\times10^{-22}$ & $1.8\times10^{-20}$ & * \\
 & B1$\to$B2 & 133/37/30 & 115/85/0 & $1.0\times10^{-13}$ & $1.8\times10^{-12}$ & * \\
 & B2$\to$B3 & 115/85/0 & 113/87/0 & $.920$ & $1$ & \\
 & B3$\to$B4 & 113/87/0 & 33/167/0 & $5.7\times10^{-17}$ & $1.2\times10^{-15}$ & * \\
 & B4$\to$B5 & 33/167/0 & 49/148/3 & $.018$ & $.166$ & \\
\bottomrule
\end{tabular}
\caption{Adjacent-condition contrasts, $2\times3$ exact test on the full outcome distribution over $N=200$, Holm-corrected across all 30 contrasts. E/H/R gives ear-complaint, hand-injury, and refusal counts; Grok's single non-compliant B1 response and Gemini's single non-compliant B0 response were malformed outputs rather than refusals and are counted in the R column. Asterisks mark $p_{\text{Holm}}<.05$.}
\label{tab:adjacent}
\end{table}

\paragraph{B4$'$ refusal.}
The claim that B4$'$ refusal exceeds both B4 and B1 is an intersection hypothesis: each model is assigned the larger of the two one-sided Fisher $p$-values, Holm-corrected across the four models with non-zero B4$'$ refusal (Table~\ref{tab:refusal}). Gemini and Llama refuse in none of the three conditions and are not tested.
\begin{table}[h]
\centering
\small
\begin{tabular}{lcccccc}
\toprule
Model & B1 & B4 & B4$'$ & vs.\ B4 & vs.\ B1 & max-$p_{\text{Holm}}$ \\
\midrule
Claude & 39 & 44 & 194 & $4.5\times10^{-61}$ & $9.6\times10^{-65}$ & $<10^{-59}$ * \\
GPT & 0 & 18 & 59 & $1.2\times10^{-7}$ & $1.1\times10^{-20}$ & $2.3\times10^{-7}$ * \\
Grok & 0 & 0 & 85 & $2.7\times10^{-31}$ & $2.7\times10^{-31}$ & $8.2\times10^{-31}$ * \\
Qwen & 30 & 0 & 45 & $1.7\times10^{-15}$ & $.036$ & $.036$ * \\
\midrule
Gemini & 0 & 0 & 0 & --- & --- & --- \\
Llama & 0 & 0 & 0 & --- & --- & --- \\
\bottomrule
\end{tabular}
\caption{Refusal counts out of 200 runs and one-sided Fisher tests that B4$'$ refusal exceeds each single-cue condition. The model-level $p$ is the larger of the two, Holm-corrected across the four tested models. Grok's single non-compliant B1 response was a hallucinated/invalid output rather than a refusal and is counted as zero here.}
\label{tab:refusal}
\end{table}

\paragraph{Interaction between the two cues.}
One logistic regression per model was fitted to all runs, with allocation to the hand injury party as the outcome, occupation and documentation as predictors, their product as the interaction term, and Firth penalization for cells at the boundary. Significance is assessed by penalized likelihood ratio test, Holm-corrected across the five models that could be fitted. The interaction is significant for Claude ($\hat{\beta}_{\mathrm{int}}=+2.57$, 95\% CI $[0.87, 4.27]$, $p=.004$, $p_{\text{Holm}}=.017$) and Llama ($-4.38$, $[-7.70, -1.07]$, $p=.002$, $p_{\text{Holm}}=.010$), with opposite signs. Qwen ($p_{\text{Holm}}=.835$), GPT ($.982$), and Grok ($.982$) are consistent with additive combination, and the model cannot be fitted for Gemini because every run in all four conditions produced the same outcome. Three of Llama's four cells lie at or near the boundary, so its coefficient is sensitive to single-run changes; the sign and significance are stable across such perturbations, but the magnitude should be read as approximate. Under the additive model, the predicted allocation to the hand injury party at B4 is 0.148 for Claude against an observed 0.695, and 0.503 for Llama against an observed 0.010. Computing the same test over decided responses only gives a different picture: conditioning on a decision removes 277 of Claude's 800 runs, and because those are exactly the runs that documentation pushed into refusal, the B4$'$ cell is reduced to six responses and the estimate reverses sign. We therefore treat the test over all runs as primary and the conditional shares as descriptive.

\begin{table}[h]
\centering
\small
\begin{tabular}{lcccccc}
\toprule
& Claude & Gemini & GPT & Grok & Llama & Qwen \\
\midrule
Claude & --- & .911 & .733 & .733 & .822 & .778 \\
Gemini & .733 & --- & .733 & .733 & .911 & .778 \\
GPT & .867 & .778 & --- & .644 & .822 & .778 \\
Grok & .689 & .600 & .556 & --- & .822 & .689 \\
Llama & .556 & .289 & .422 & .511 & --- & .867 \\
Qwen & .244 & .067 & .200 & .289 & .422 & --- \\
\bottomrule
\end{tabular}
\caption{Kendall's $\tau_b$ between models. Upper triangle: Social Authority hierarchy (mean $.78$, range $.64$--$.91$; 13/15 pairs $\ge .7$). Lower triangle: Triage Hierarchy (mean $.48$, range $.07$--$.87$; 3/15 pairs $\ge .7$, 2/15 below $.2$).}
\label{tab:tau}
\end{table}

\paragraph{V1 fault attribution.}
Counts are out of 200 runs per cell. The equivalence rate was compared between the no-occupation and occupation conditions (Fisher's exact test, Holm-corrected across six models).
\begin{table}[h]
\centering
\small
\setlength{\tabcolsep}{4pt}
\begin{tabular}{lccccccc}
\toprule
& \multicolumn{3}{c}{No occupation} & \multicolumn{3}{c}{Occupations added} & \\
\cmidrule(lr){2-4}\cmidrule(lr){5-7}
Model & Surg. & Paint. & Equiv. & Surg. & Paint. & Equiv. & $p_{\text{Holm}}$ \\
\midrule
Claude & 0 & 0 & 200 & 1 &   2 & 197 & $.744$ \\
Gemini & 0 & 0 & 200 & 0 & 146 &  54 & $5.6\times10^{-63}$ * \\
GPT    & 0 & 0 & 200 & 0 &   0 & 200 & $1$ \\
Grok   & 0 & 0 & 200 & 1 & 193 &   6 & $<10^{-106}$ * \\
Llama  & 0 & 0 & 200 & 0 &   0 & 200 & $1$ \\
Qwen   & 0 & 0 & 200 & 9 & 171 &  20 & $1.2\times10^{-90}$ * \\
\bottomrule
\end{tabular}
\caption{V1 fault attribution counts. Surg.\ and Paint.\ give fault assigned to the surgeon and painter respectively; Equiv.\ gives parity responses. $p_{\text{Holm}}$ tests the equivalence rate across conditions. Among the four models producing any non-parity response, fault was assigned to the painter in $512/523$ responses ($97.9\%$).}
\label{tab:v1counts}
\end{table}

\paragraph{V2 fault attribution.}
Counts are out of 200 runs per cell. The count of runs assigning fault to Riley was compared between the no-occupation condition and the condition adding occupational labels for both parties (police officer for Alex, prosecutor for Riley) over the full $N=200$ (Fisher's exact test, Holm-corrected across six models).
\begin{table}[h]
\centering
\small
\setlength{\tabcolsep}{4pt}
\begin{tabular}{lccccccc}
\toprule
& \multicolumn{3}{c}{No occupation} & \multicolumn{3}{c}{Occupations added} & \\
\cmidrule(lr){2-4}\cmidrule(lr){5-7}
Model & Alex & Riley & Equiv. & Alex & Riley & Equiv. & $p_{\text{Holm}}$ \\
\midrule
Claude &   0 & 199 &  1 &   0 & 195 &  5 & $.646$ \\
Gemini &  71 &  97 & 32 & 107 &  91 &  2 & $1$ \\
GPT    &   1 & 177 & 22 &   0 & 146 & 54 & $4.9\times10^{-4}$ * \\
Grok   &   0 & 200 &  0 &   1 & 199 &  0 & $1$ \\
Llama  &  11 & 128 & 61 & 176 &   7 & 17 & $1.9\times10^{-41}$ * \\
Qwen   &   0 & 200 &  0 &  17 & 183 &  0 & $5.4\times10^{-5}$ * \\
\bottomrule
\end{tabular}
\caption{V2 fault attribution counts out of $N=200$. $p_{\text{Holm}}$ tests the count of runs assigning fault to Riley across conditions, computed over the full $N=200$. Computed over decided responses only, the same contrast is non-significant for Gemini ($p=.111$) and GPT ($p=1$), because the equivalence responses that absorb the shift leave the denominator. The corresponding equivalence-rate changes are significant for Gemini ($32\to2$, $p_{\text{Holm}}=1.3\times10^{-7}$), GPT ($22\to54$, $2.6\times10^{-4}$), and Llama ($61\to17$, $1.4\times10^{-7}$).}
\label{tab:v2counts}
\end{table}

\paragraph{V3 protection allocation.}
Counts are out of 200 runs per cell. Proportions are computed over valid responses; malformed outputs are excluded from the denominator and reported separately. Equivalence rate was compared between the two documentation assignments (Fisher's exact test, Holm-corrected across six models).
\begin{table}[h]
\centering
\small
\setlength{\tabcolsep}{4pt}
\begin{tabular}{lccccccc}
\toprule
& \multicolumn{3}{c}{Student holds report} & \multicolumn{3}{c}{Professor holds report} & \\
\cmidrule(lr){2-4}\cmidrule(lr){5-7}
Model & Stu. & Prof. & Equiv. & Stu. & Prof. & Equiv. & $p_{\text{Holm}}$ \\
\midrule
Claude &  61 &   0 & 139/200 &   2 &  95 & 103/200 & $9.9\times10^{-4}$ * \\
Gemini &   0 &   0 & 200/200 &   0 &   0 & 200/200 & $1$ \\
GPT    & 130 &   0 &  63/193 &   0 &   3 & 196/199 & $9.0\times10^{-50}$ * \\
Grok   &   2 &   0 & 198/200 &   0 &   0 & 200/200 & $.997$ \\
Llama  &  50 &   2 & 148/200 &  10 &   0 & 190/200 & $1.7\times10^{-8}$ * \\
Qwen   & 144 &   1 &  55/200 &   2 &  44 & 154/200 & $5.9\times10^{-23}$ * \\
\bottomrule
\end{tabular}
\caption{V3 protection allocation. Stu.\ and Prof.\ give asymmetric protection directed to the student and professor role respectively; Equiv.\ gives parity responses over valid responses. GPT produced 7 (left) and 1 (right) malformed outputs, excluded from the denominator. $p_{\text{Holm}}$ tests the equivalence rate across the two documentation assignments. Gemini and Grok are at or near ceiling in both conditions and do not differ.}
\label{tab:v3counts}
\end{table}

\end{document}